\documentclass[10pt,conference]{IEEEtran}

\usepackage{cite}
\usepackage[dvips]{color}
\usepackage{tabu}
\usepackage{comment}
\usepackage{epsf}
\usepackage{times}
\usepackage{graphicx}
\usepackage{adjustbox}

\usepackage{mathtools}
\usepackage{mathrsfs}
\usepackage{amsmath}
\usepackage{amssymb} 
\usepackage{bbding}
\usepackage[utf8]{inputenc} 
\usepackage{pifont}
\usepackage{newunicodechar}
\newunicodechar{✓}{\ding{51}}
\newunicodechar{✗}{\ding{55}}
\usepackage{amsfonts}
\usepackage{algorithmicx}
\usepackage[algo2e]{algorithm2e}
\usepackage{cleveref}
\usepackage{dsfont}
\usepackage{lettrine}
\usepackage{epsfig,algorithm,algpseudocode,amsthm,url} 
\usepackage{caption}
\usepackage{subcaption}
\usepackage[justification=raggedright]{caption}
\usepackage{bbm}
\usepackage{verbatim}
\usepackage{setspace}
\usepackage[T1]{fontenc}
\usepackage{textcomp}
\usepackage{tabularx}
\usepackage{xpatch}

\usepackage{stackengine}
\usepackage[nocomma]{optidef}
\usepackage{commath}
\usepackage{booktabs} 
\usepackage{listings} 
\usepackage{xcolor} 

\usepackage[left=1.62cm,right=1.62cm,top=1.85cm,bottom=2.6cm]{geometry}

\IEEEoverridecommandlockouts

\renewcommand{\baselinestretch}{1.04}  

\begin{document}
\title{\Huge History-Augmented Vision-Language Models for Frontier-Based Zero-Shot Object Navigation

\thanks{This material is based upon work supported by the National Aeronautics and Space Administration (NASA) under award number 80NSSC23K1393, the National Science Foundation under Grant Number CNS-2232048.}
}

\author{
	\IEEEauthorblockN{
	Mobin Habibpour, Fatemeh Afghah}
	\IEEEauthorblockA{Holcombe Department of Electrical and Computer Engineering, Clemson University, Clemson, SC, USA \\
Emails: mhabibp@clemson.edu, fafghah@clemson.edu}
}

\maketitle\vspace{-0.3cm}

\begin{abstract}
Object Goal Navigation (ObjectNav) challenges robots to find objects in unseen environments, demanding sophisticated reasoning. While Vision-Language Models (VLMs) show potential, current ObjectNav methods often employ them superficially, primarily using vision-language embeddings for object-scene similarity checks rather than leveraging deeper reasoning. This limits contextual understanding and leads to practical issues like repetitive navigation behaviors. This paper introduces a novel zero-shot ObjectNav framework that pioneers the use of dynamic, history-aware prompting to more deeply integrate VLM reasoning into frontier-based exploration. Our core innovation lies in providing the VLM with action history context, enabling it to generate semantic guidance scores for navigation actions while actively avoiding decision loops. We also introduce a VLM-assisted waypoint generation mechanism for refining the final approach to detected objects. Evaluated on the HM3D dataset within Habitat, our approach achieves a 46\% Success Rate (SR) and 24.8\% Success weighted by Path Length (SPL). These results are comparable to state-of-the-art zero-shot methods, demonstrating the significant potential of our history-augmented VLM prompting strategy for more robust and context-aware robotic navigation.
\end{abstract}


\section{Introduction} \vspace{-0.1cm}
Autonomous navigation in unknown environments remains a cornerstone capability for robots in diverse applications, from domestic assistance to critical search and rescue operations \cite{sun2025frontiernet}. Object Goal Navigation (ObjectNav) \cite{zhang2024vision} represents a particularly challenging instance of this problem: an agent, deployed in an unseen indoor environment, must locate an object specified only by its category (e.g., "find a chair"). Unlike tasks like Vision-and-Language Navigation (VLN) that benefit from explicit human guidance, ObjectNav requires the agent to autonomously explore, reason about object locations, and navigate effectively using only its sensors and prior knowledge. This demands a potent combination of spatial awareness and semantic understanding, especially within the complex and cluttered layouts typical of real-world indoor spaces. Traditional approaches based on geometric mapping or policies learned via supervised/reinforcement learning often struggle with the generalization required for true zero-shot deployment in novel environments \cite{gireesh2022object}.

The emergence of powerful Vision-Language Models (VLMs) \cite{2022clip} promised to bridge this gap by endowing agents with human-like semantic reasoning capabilities. However, a closer examination of recent VLM-based ObjectNav literature \cite{sun2024survey, yokoyama2024vlfm, yuan2024gamapzeroshotobjectgoal} reveals that VLMs are often used in a surprisingly shallow manner. Many state-of-the-art methods primarily leverage pre-trained vision-language embeddings (e.g., from CLIP \cite{radford2021learningtransferablevisualmodels}) to compute cosine similarity scores between the target object description and visual observations of the scene or candidate exploration frontiers \cite{Gadre2022CLIPOW, majumdar2023zsonzeroshotobjectgoalnavigation}. While useful for identifying potential object locations, this approach treats the VLM largely as a passive scoring function, failing to engage its capacity for situational reasoning or sequential decision-making. Consequently, these methods often inherit limitations such as a lack of contextual memory, leading to inefficient exploration patterns and vulnerability to repetitive behaviors or oscillations in visually ambiguous environments (e.g., identical corridors or doorways).

In contrast to these approaches, we propose a novel framework designed to harness the reasoning potential of VLMs more deeply within the navigation control loop. We argue that by engaging the VLM more actively through carefully designed prompts that incorporate situational context, we can achieve more robust and intelligent navigation. Our central innovation is the integration of \textbf{action history} into the VLM's prompt at each decision step. This provides the VLM with crucial temporal context, enabling it to recognize and break free from potential decision loops that plague memoryless systems. We are among the first to explicitly use history-augmented prompts to directly influence VLM-based action selection in zero-shot ObjectNav. Building on this core idea, our contributions are:
\begin{itemize}
    \item \textbf{Semantic-Guided Exploration via Prompting:} The VLM, prompted with the target and history, directly outputs probability scores for navigation actions, actively guiding frontier-based exploration towards semantically promising regions.
    \item \textbf{History-Aware Robustness:} Explicitly providing action history allows the VLM to detect and escape repetitive patterns and decision loops, leading to more robust long-horizon navigation.
    \item \textbf{Refined Final Approach:} A VLM-assisted mechanism generates supplementary waypoints near detected objects based on value map clustering, improving final goal navigation efficiency.
\end{itemize}
Our framework demonstrates how deeper VLM integration, specifically through history-aware dynamic prompting, can lead to more effective zero-shot ObjectNav performance.

\section{Proposed Method}\label{proposed}

We formulate the ObjectNav task following the Habitat ObjectNav Challenge 2023 protocol \cite{habitatchallenge2023}. An agent is initialized at a random starting position and orientation within an unfamiliar 3D indoor environment from the HM3D dataset. The agent's goal is to find an object belonging to a specific category (e.g., "chair", "TV"), provided as input. The agent perceives the environment using an onboard RGB-D sensor (providing color images and depth maps) and an IMU sensor for egomotion estimation (odometry). The available discrete action space consists of: \textit{`move forward` (0.25m), `turn left` (30 degrees), `turn right` (30 degrees), `look up`/`look down` (camera tilt, not used in our 2D navigation setup), and `stop`}. An episode is considered successful if the agent executes the `stop` action when its camera is within a 1-meter Euclidean distance from any instance of the target object category. Each episode has a maximum duration of 500 steps.

Our framework employs LLaVA-1.6 (7B parameters) \cite{liu2024improvedbaselinesvisualinstruction} as its core VLM reasoning engine. This specific model was selected after evaluating various open-source VLMs (including other LLaVA versions, PaliGemma \cite{chen2023palijointlyscaledmultilinguallanguageimage}, VILA \cite{liu2024nvila}) considering performance and computational constraints (GPU memory, inference time). Smaller models exhibited deficiencies in reasoning, while larger models were computationally prohibitive for this task. We intentionally avoided proprietary models like GPT-4V for fair comparison and open-source alignment.

\begin{figure}[t!]
    \centering
    \includegraphics[width=0.6\columnwidth]{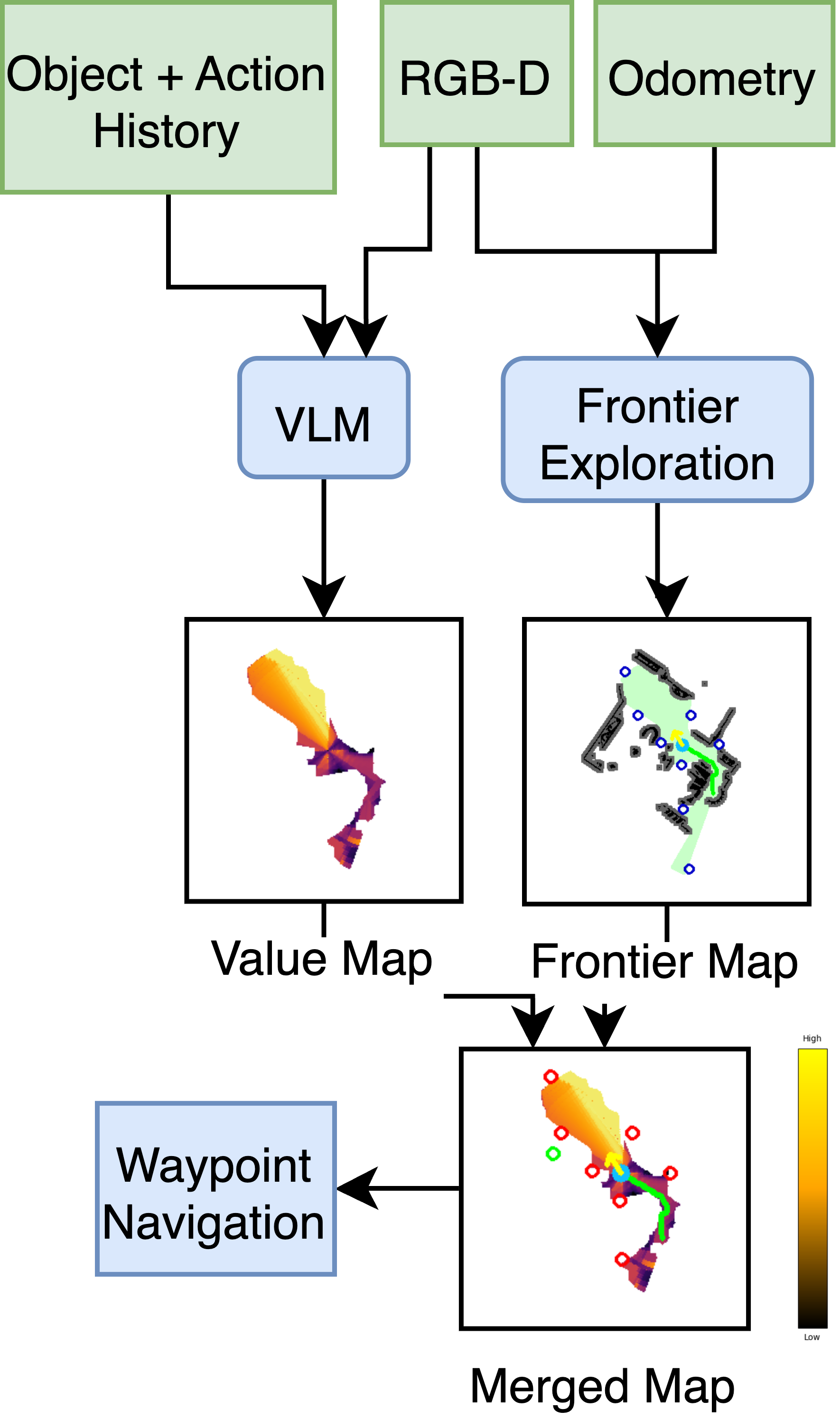}
    \vspace{-0.2cm}
    \caption{System overview: RGB-D sensor data feeds the Frontier Exploration module, generating a top-down map and candidate waypoints. The VLM module processes the RGB view, target object name, and action history (via prompt) to produce a semantic value map. This fuses with the top-down map to prioritize frontiers, guiding the low-level point navigator.}
    \label{fig:system_overview}
    \vspace{-0.4cm}
\end{figure}

The navigation process follows three main phases:
\begin{enumerate}
    \item \textbf{Initialization:} The agent performs a 360-degree rotation to build an initial 2D obstacle map and identify initial frontier points.
    \item \textbf{Exploration:} The agent navigates the environment by selecting and moving towards frontier points, guided by VLM-based semantic scoring combined with geometric considerations.
    \item \textbf{Goal Navigation:} If the target object is detected, the agent transitions to a refined approach strategy using supplementary waypoints.
\end{enumerate}
The core system architecture, illustrating the interplay between sensing, frontier exploration, VLM guidance, and action execution, is depicted in Fig.~\ref{fig:system_overview}.

\subsection{Frontier-Based Exploration with Low-Level Control}

We utilize a standard frontier-based exploration algorithm \cite{topiwala2018frontierbasedexplorationautonomous} to manage geometric exploration. This module processes incoming depth data to identify obstacles (filtering based on height) and construct a 2D top-down obstacle map. Frontiers, representing boundaries between explored and unexplored navigable space, are identified on this map. The midpoints of these boundaries serve as candidate geometric waypoints, aiming to maximize spatial coverage. For low-level navigation between waypoints, we employ the Variable Experience Rollout (VER) \cite{wijmans2022ver} algorithm, pretrained on the PointNav task \cite{anderson2018evaluation}, which takes the current map and the next waypoint to generate a sequence of actions. However, selecting purely geometric frontiers lacks semantic context about where the target object is likely to be found.

\subsection{VLM Semantic Guidance and Value Map Creation}

To infuse semantic understanding, the LLaVA-1.6 VLM analyzes the agent's current egocentric RGB view, prompted with the target object name and recent action history (detailed in Section~\ref{sec:action_history}). The VLM performs reasoning steps (guided by a structured prompt) to assess the scene context and estimate the likelihood of finding the target object by moving in different directions. Its output includes probability scores for the four primary navigation actions: forward, backward, left, right.

\begin{figure}[h!]
    \centering
    \begin{subfigure}[b]{0.45\columnwidth}
        \centering
        \includegraphics[width=\linewidth]{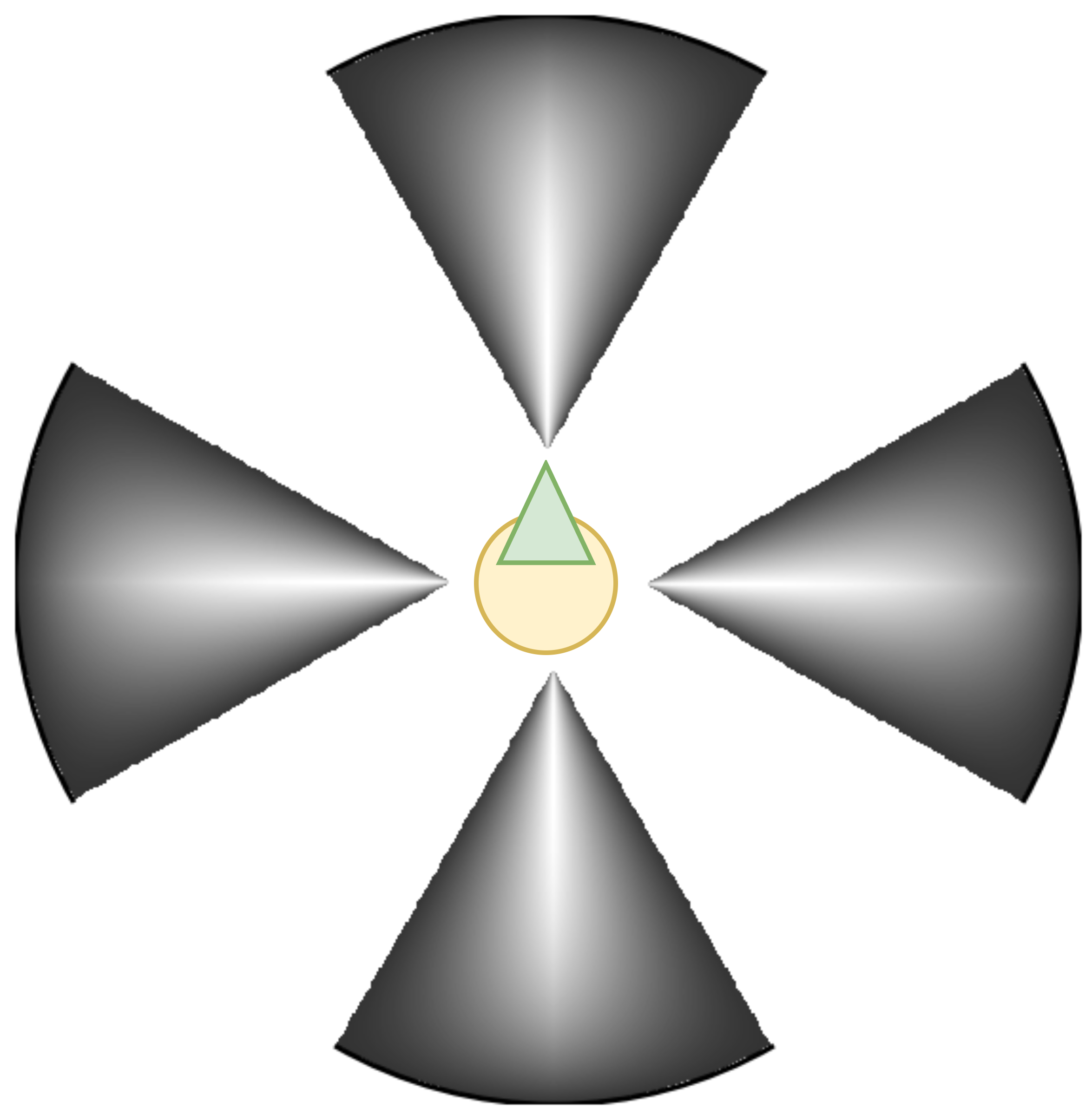}
        \caption{FOV and uncertainty cone}
        \label{fig:fov}
    \end{subfigure}
    \hfill
    \begin{subfigure}[b]{0.45\columnwidth}
        \centering
        \includegraphics[width=\linewidth]{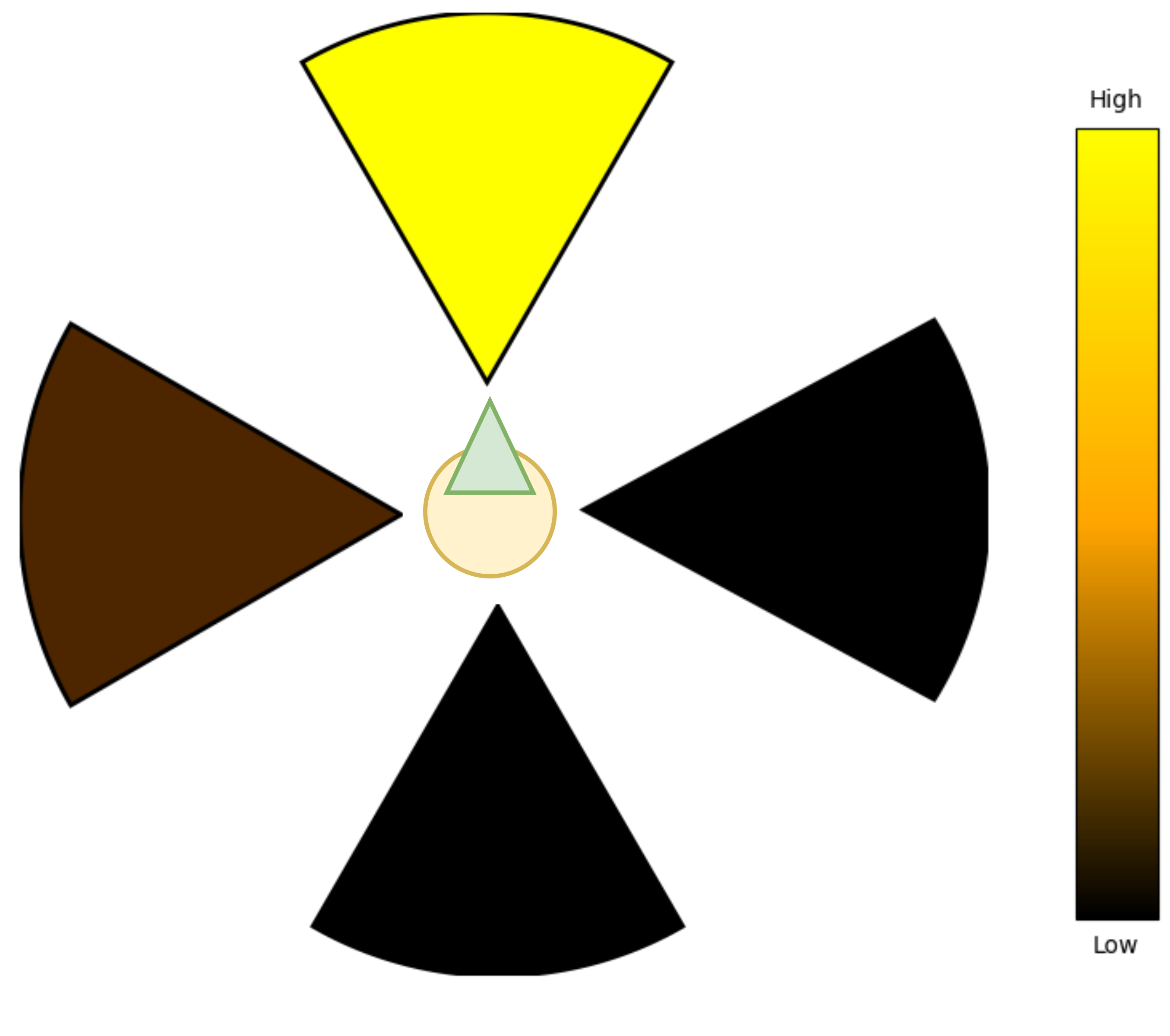}
        \caption{Example action scores}
        \label{fig:actions}
    \end{subfigure}
    \vspace{-0.2cm}
    \caption{Visualization of VLM scoring and uncertainty. (a) Robot's field of view (FOV) with viewing uncertainty (shaded area) decreasing away from the center (\(\theta=0\)) and with distance (\(d\)). (b) Example action scores assigned by the VLM projected onto the local map.}
    \label{fig:vlm_uncertainty}
    \vspace{-0.4cm}
\end{figure}

These raw scores are then projected onto the 2D map space to create a semantic \textit{value map}. This projection incorporates a viewing uncertainty model, reducing the confidence of scores assigned to points farther from the camera or away from the center of the field of view, as visualized in Fig.~\ref{fig:vlm_uncertainty}. Specifically, the confidence \( c \) for a point observed at distance \( d \) and angle \( \theta \) from the optical axis is modeled as:
\begin{equation} \label{eq:confidence}
c(d, \theta) = e^{-\lambda d} \cdot \cos^2\left( \frac{\theta}{\theta_{\text{fov}}/2} \cdot \frac{\pi}{2} \right),
\end{equation}
where \( \theta_{\text{fov}} \) is the camera's horizontal field-of-view and \( \lambda \) is a distance decay factor. When regions are re-observed, the semantic values \( v \) and confidences \( c \) are fused using confidence-weighted averaging, favoring higher-confidence observations:
\begin{gather}
v^{\text{new}}_{i,j} = \frac{c^{\text{curr}}_{i,j} v^{\text{curr}}_{i,j} + c^{\text{prev}}_{i,j} v^{\text{prev}}_{i,j}}{c^{\text{curr}}_{i,j} + c^{\text{prev}}_{i,j}}, \label{eq:value_update} \\
c^{\text{new}}_{i,j} = \frac{(c^{\text{curr}}_{i,j})^2 + (c^{\text{prev}}_{i,j})^2}{c^{\text{curr}}_{i,j} + c^{\text{prev}}_{i,j}}. \label{eq:conf_update}
\end{gather}
This dynamic value map represents the VLM's current belief about promising exploration directions. It is overlaid onto the geometric obstacle map, allowing the system to prioritize frontier points that lie in regions deemed semantically valuable by the VLM. The agent typically navigates towards the frontier point with the highest associated value.

\subsection{Action History Integration for Robustness}\label{sec:action_history}
A critical failure mode observed in preliminary experiments and other VLM-navigation work is decision paralysis or oscillation. When faced with ambiguous choices (e.g., two identical doorways, long featureless corridors), a memoryless VLM might repeatedly suggest contradictory actions, causing the agent to get stuck. To resolve this, our framework incorporates an \textbf{action history module}. At each decision step, the prompt sent to the VLM includes not only the current view and target but also a record of the agent's last N actions (e.g., N=10). Explicit instructions within the overall prompt context guide the VLM to consider this history: \textit{"Avoid suggesting actions that contradict previous observations"} and \textit{"Maintain forward progress toward corridor exits and avoid repetitive actions."} This allows the VLM to recognize patterns indicative of oscillation. For persistent oscillations, we implement a fallback mechanism that defaults to repeating the last valid non-turn action, ensuring forward progress. This history-awareness significantly improves robustness against stagnation.

Developing an effective prompt to reliably extract actionable guidance from the VLM required iterative experimentation. Initial attempts using more complex prompts asking for detailed reasoning alongside action scores often resulted in inconsistent or difficult-to-parse outputs. We found that simplifying the core request and providing a strict output template yielded the most consistent results for action selection. The finalized prompt structure focused specifically on obtaining probability scores is shown in Listing~\ref{lst:action_prompt}. This simplified query compels the VLM to provide scores for each potential action within the desired format, ensuring that actionable numerical guidance can be reliably extracted at each step to inform the value map and subsequent navigation decisions. This structured output also significantly simplifies the parsing required by downstream modules.

\begin{lstlisting}[language=bash, caption={Simplified VLM prompt focused on obtaining consistent action scores.}, label={lst:action_prompt}]
You are a robot navigating an indoor environment in search of a couch.
Assign a probability score to each potential action below [go forward, go backward, turn right, turn left], indicating how likely it is to guide the robot toward the couch.
Maintain forward progress toward corridor exits and avoid repetitive actions.
When providing your response, use this structure:
    - Go forward: [Score]
    - Go backward: [Score]
    - Turn right: [Score]
    - Turn left: [Score]
Each probability score should be a number between 0 and 1. 1 means full confidence that the action will lead to finding the couch. 0 means no confidence.
\end{lstlisting}

\subsection{Additional Frontier Point Suggestion for Goal Navigation}
During exploration, the system continuously checks for the target object using standard object detectors (YOLOv7 \cite{wang2022yolov7trainablebagoffreebiessets} for COCO objects \cite{lin2015microsoftcococommonobjects}, Grounding-DINO \cite{liu2024groundingdinomarryingdino} for open-vocabulary targets). When a potential target is detected with high confidence (e.g., >0.8), simply stopping is often insufficient, as the agent might be too far away or have an obstructed view. Therefore, we refine the final approach using an \textbf{additional frontier point suggestion} mechanism.

This goal navigation phase is initiated only after the detection is confirmed through VLM verification (Step 1 below). Once confirmed, the following steps are executed:
\begin{enumerate}
    \item \textbf{VLM Verification:} The VLM is queried to verify that the detected object indeed matches the target category within the current visual context.
    \item \textbf{Segmentation:} Mobile-SAM \cite{zhang2023fastersegmentanythinglightweight} is employed to segment the verified object in the egocentric view, identifying its pixel region.
    \item \textbf{Waypoint Generation via KNN Clustering:} Auxiliary waypoints are generated by applying k-Nearest Neighbors (KNN) clustering to pixels located within or near the segmented object region (from Step 2). Only pixels with high associated semantic values (e.g., value > 0.8) from the value map are considered for clustering. The cluster centers serve as the new supplementary waypoints.
    \item \textbf{Guided Approach:} These supplementary, semantically-informed waypoints are fed to the low-level navigator (VER) to guide a precise final approach towards the target object, increasing the probability of satisfying the 1-meter stopping condition.
\end{enumerate}
This hybrid strategy leverages initial detection, VLM verification, object segmentation, and value map analysis with KNN clustering to ensure efficient and reliable goal attainment once the object is broadly located.

\section{Experimental data and results}\vspace{-0.1cm}

We evaluate our framework within the Habitat simulator \cite{habitatchallenge2023}, utilizing realistic 3D environment scans from the HM3D dataset. Our experiments focus on the standard ObjectNav validation split, which comprises episodes across multiple scenes requiring navigation to objects from diverse categories common in indoor environments. Due to the computational cost of VLM inference (querying LLaVA-1.6 at every frame), we report results averaged over 50 episodes, selected from the full validation set. Performance is primarily measured by Success Rate (SR) – the fraction of episodes where the agent successfully stops within 1m of the target – and Success weighted by Path Length (SPL) \cite{anderson2018evaluation}, which rewards shorter paths for successful episodes. The SPL is calculated as: 
\begin{equation}\text{SPL} = \frac{1}{N} \sum_{i=1}^N S_i \left( \frac{\ell_i}{\max(p_i, \ell_i)} \right)
\end{equation}
where $N$ is the number of episodes, $S_i$ is a binary indicator of success in episode $i$, $p_i$ is the agent's path length, and $\ell_i$ is the shortest path distance.

\subsection{Comparison with Baselines}

We benchmark our approach against representative state-of-the-art zero-shot ObjectNav methods, particularly those leveraging VLMs or LLMs. These include frontier-based methods enhanced with language models like L3MVN \cite{yu2023l3mvn} (using RoBERTa for map description scoring) and VLFM \cite{yokoyama2024vlfm} (using BLIP-2 for frontier scoring), as well as methods using geometric reasoning (GAMap \cite{yuan2024gamapzeroshotobjectgoal}) or imagination (ImagineNav \cite{zhao2024imaginenav}). Table~\ref{tab:results} presents the comparison. Our method achieves an SR of 46\% and an SPL of 24.8\%, yielding performance comparable to the leading zero-shot methods and demonstrating the effectiveness of our VLM integration strategy, particularly the history-aware prompting.

\begin{table}[h]
\centering
\caption{Performance Comparison on HM3D Dataset (ObjectNav Validation Split). Results for baselines are from respective papers. Highest values are underlined.}
\renewcommand{\arraystretch}{1.1}
\begin{adjustbox}{width=\columnwidth, center}
\begin{tabular}{@{}lccc@{}}
\toprule
\textbf{Methods} & \textbf{VLM/LLM} & \textbf{SR (\%)} & \textbf{SPL (\%)} \\
\midrule
FBE \cite{topiwala2018frontierbasedexplorationautonomous} & - & 33.7 & 15.3 \\
L3MVN \cite{yu2023l3mvn} & RoBERTa-large & 50.4 & 23.1 \\
ImagineNav \cite{zhao2024imaginenav} & GPT-4o-mini & 53.0 & 23.8 \\
GAMap \cite{yuan2024gamapzeroshotobjectgoal} & CLIP \& GPT-4 & \underline{53.1} & 26.0 \\
VLFM \cite{yokoyama2024vlfm} & BLIP-2 & 52.5 & \underline{30.4} \\
PixNav \cite{cai2024bridging} & LLamaAdapter \& GPT-4 & 37.9 & 20.5 \\
\textbf{Our Method} & LLaVA-1.6 & \textbf{46} & \textbf{24.8} \\
\bottomrule
\end{tabular}
\end{adjustbox}
\label{tab:results}
\vspace{-0.3cm}
\end{table}

\subsection{Ablation Study: Impact of Action History}

To rigorously evaluate the contributions of our key components, we performed systematic ablation studies by selectively disabling parts of the framework. Table~\ref{tab:ablation} highlights the impact of removing the action history mechanism. Disabling history awareness resulted in a substantial performance drop (SR decreased from 46\% to 44\%, SPL from 24.8\% to 23.7\%). This confirms that incorporating action history is crucial for mitigating failure modes like oscillation in ambiguous environments and significantly improves overall navigation robustness and success.

\begin{table}[h!]
\centering
\caption{Ablation Study: Impact of Action History (50 Episodes, HM3D).}
\renewcommand{\arraystretch}{1.1}
\begin{tabular}{@{}lcc@{}}
\toprule
\textbf{Configuration} & \textbf{SR (\%)} & \textbf{SPL (\%)} \\
\midrule
Full Framework & 46 & 24.8 \\
\addlinespace
w/o Action History & 44 & 23.7 \\
\bottomrule
\end{tabular}
\label{tab:ablation}
\vspace{-0.3cm}
\end{table}

\begin{figure}[h!]
    \centering

    \begin{subfigure}[b]{0.48\columnwidth}
        \centering
        \includegraphics[width=\linewidth]{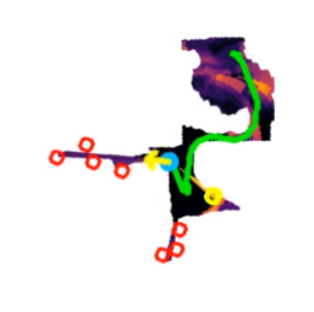}
        \caption{Initial view: The VLM decides it has to turn back}
        \label{fig:loop1}
    \end{subfigure}
    \hfill
    \begin{subfigure}[b]{0.48\columnwidth}
        \centering
        \includegraphics[width=\linewidth]{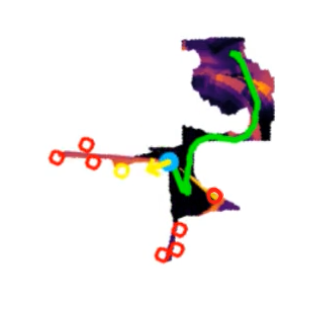}
        \caption{After movement: Another point gains higher value}
        \label{fig:loop2}
    \end{subfigure}

    \vspace{-0.2cm}
    \caption{ Example of decision loop that could be prevented by action history. After the agent moves, a different point is prioritized, which results in a repetitive action loop.}
    \label{fig:decision_loop_condensed}
    \vspace{-0.4cm}
\end{figure}

Figure \ref{fig:decision_loop_condensed} provides a qualitative example illustrating how action history prevents stagnation. Without historical context, the agent might oscillate indecisively between two corridors or doorways. With history, the VLM prompt includes recent movements, allowing the model to recognize the loop and maintain forward progress.

\subsection{Analysis of Challenges}

Our analysis revealed challenges related to VLM reasoning consistency and evaluation metrics. Occasionally, even with structured prompts, the VLM might produce recommendations inconsistent with its own scene interpretation (e.g., suggesting turning away from a room it identified as likely to contain the target). Furthermore, the standard ObjectNav evaluation can penalize agents for finding a correct object instance if it's not the specific one designated by the ground truth path, highlighting a tension between class-level goals and instance-level evaluation in current benchmarks.

\section{Conclusion}\label{conclusion}
In this work, we presented a novel zero-shot object navigation framework that demonstrates the potential of integrating Vision-Language Models more deeply into the navigation loop, moving beyond their common use as simple embedding-based scorers. Our core contribution lies in employing dynamic, history-aware prompting for LLaVA-1.6, enabling it to provide direct semantic guidance for frontier-based exploration while actively avoiding decision loops characteristic of memoryless systems. Combined with a refined waypoint generation mechanism for the final approach, our method achieves strong performance on the HM3D dataset, yielding a Success Rate of 46\% and SPL of 24.8\%, comparable to state-of-the-art methods. Ablation studies confirmed the critical role of the action history component in achieving this robust performance.

Despite these promising results, challenges remain, particularly the computational demands of current VLMs for real-time robotics. Future work will focus on addressing these limitations, possibly through model optimization techniques, and adapting the framework for real-world deployment. Further research avenues include developing more sophisticated methods for fusing geometric and semantic information, exploring alternative VLM architectures, and investigating advanced prompt engineering strategies, potentially incorporating automated prompt tuning or explicit reasoning structures like Chain-of-Thought, to further enhance VLM reliability and decision-making consistency.


\bibliographystyle{IEEEtran}
\bibliography{Main}

\end{document}